\documentclass[runningheads]{llncs}
\usepackage[T1]{fontenc}
\usepackage{graphicx}
\usepackage{tabularx}
\usepackage{booktabs}
\usepackage[table]{xcolor}
\usepackage{xcolor}
\usepackage{algorithm}
\usepackage{algorithmicx}
\usepackage{algpseudocode}
\usepackage{amsmath,amssymb,amsfonts}%
\usepackage{amsthm}
\usepackage{listings}
\usepackage{subfloat}
\usepackage{subcaption}
\usepackage{hyperref}
\usepackage[inline]{enumitem}
\definecolor{codegreen}{rgb}{0,0.6,0}
\definecolor{codegray}{rgb}{0.5,0.5,0.5}
\definecolor{codepurple}{rgb}{0.58,0,0.82}
\definecolor{backcolour}{rgb}{0.95,0.95,0.92}

\lstdefinestyle{mystyle}{
    backgroundcolor=\color{backcolour},   
    commentstyle=\color{codegreen},
    keywordstyle=\color{blue},
    numberstyle=\tiny\color{codegray},
    stringstyle=\color{codepurple},
    basicstyle=\ttfamily\footnotesize,
    breakatwhitespace=false,         
    breaklines=true,                 
    captionpos=b,                    
    keepspaces=true,                 
    numbers=left,                    
    numbersep=5pt,                  
    showspaces=false,                
    showstringspaces=false,
    showtabs=false,                  
    tabsize=2,
    floatplacement=h
}
\usepackage{todonotes}

\begin{document}
%
\title{Virtue Ethics For Ethically Tunable Robotic Assistants}
%
%
\author{Rajitha Ramanayake\orcidID{0000-0001-9903-0493} \and
Vivek Nallur\orcidID{0000-0003-0447-4150}}
\authorrunning{Ramanayake R. and Nallur V.}
%
\institute{University College Dublin, Belfield, Dublin. \\
\email{rajitha.ramanayakemahantha@ucdconnect.ie}, \email{vivek.nallur@ucd.ie}
}
\maketitle              
\begin{abstract}
The common consensus is that robots designed to work alongside or serve humans must adhere to the ethical standards of their operational environment. To achieve this, several methods based on established ethical theories have been suggested. Nonetheless, numerous empirical studies show that the ethical requirements of the real world are very diverse and can change rapidly from region to region. This eliminates the idea of a universal robot that can fit into any ethical context. However, creating customised robots for each deployment, using existing techniques is challenging. This paper presents a way to overcome this challenge by introducing a virtue ethics inspired computational method that enables character-based tuning of robots to accommodate the specific ethical needs of an environment. Using a simulated elder-care environment, we illustrate how tuning can be used to change the behaviour of a robot that interacts with an elderly resident in an ambient-assisted environment. Further, we assess the robot's responses by consulting ethicists to identify potential shortcomings.

\end{abstract}
\section{Introduction}
Ethical behaviour represents a crucial aspect of Human-Robot Interaction (HRI) in real-world applications. It is commonly agreed that the robots that share their working environments with humans, such as those that provide care, assistance, transportation, or companionship to humans, must adhere to and reason about their behaviour within the ethical framework of the community around them~\cite{Wallach2008}. This is especially true in the care domain as care robots interact with vulnerable populations that require special care and protection~\cite{Sharkey2012}. Consequently, many ways have been proposed to implement ethical behaviour in robots. Many proposed techniques adopt a top-down approach, based on ethical theories such as deontology~\cite{sep-ethics-deontological} or consequentialism~\cite{sep-consequentialism} (e.g.,~\cite{benzmuller_designing_2020,van_dang_application_2017,Vanderelst2018a}). A relatively small number of proposed methods use a bottom-up approach where they utilise learning algorithms such as reinforcement learning, that enable the robots to learn the ethical behaviours from the environment (e.g.,~\cite{solaiman_process_2021,Abel2016}). Another set of implementations uses a hybrid approach where they integrate both top-down and bottom-up methods in different parts of their ethical framework (e.g.,~\cite{Anderson2018,Azad-Manjiri2014,yilmaz_computational_2017}). 

Studies have shown that the ethical requirements of the real world vary significantly, with subtle variations within different areas of the same region~\cite{Awad2018}. A potential solution to this challenge is to design tailored ethical governors for robots that suit each deployment scenario. 

A possible way to address this challenge is to implement value-oriented moral character in autonomous machines. This idea has been theoretically explored in literature~\cite{gamez_artificial_2020,pirni_robot_2021,gibert_case_2023}. This paper introduces an implementation of an ethical reasoning architecture inspired by virtue ethics. It enables character-based tuning for a system to align with the ethical demands of its environment. We demonstrate this capability through a virtual simulation of an assistive elder-care robot confronted with an ethically charged scenario. 

The rest of the paper is organised as follows. Section~\ref{related_work} discusses existing implementations of ethics in robots and the need for ethical tuning. Section~\ref{character_in_robot} discusses the underlying philosophical aspects of the notion of character in robots, and how it can be used in ethical tuning. Section~\ref{implementation} presents the computational architecture used in the implementation and briefly discusses the simulation environment. Section~\ref{results} outlines the design of the experiment, analyses the behaviour observed in the simulations, and presents the evaluation of this behaviour by ethicists. Section~\ref{discussion} discusses results and the recommendations for real-world implementations of this computational architecture.


\section{Current Approaches and Ethical Tuning} \label{related_work}


The field of Machine Ethics acknowledges that human value alignment is an important part of any robotic application that interfaces with humans in social contexts~\cite{russell_human_2019}. 
Some researchers argue that there is no value in developing explicitly ethical agents~\cite{VanWynsberghe2018}, and many others disagree~\cite{gabriel_artificial_2020,van_wynsberghe_social_2022}. 
Large-scale human preference studies such as `The moral machine experiment'~\cite{Awad2018} bring forth an interesting finding, which is that the expected behaviour of a robot, even for a minimal ethical dilemma, changes throughout the world according to the culture and the environment the robot is situated in. This is also discussed in the elder-care robots domain. Specifically, the behavioural expectations of the same robot in ethically charged scenarios can vary depending on the context in which it is used~\cite{hakli_small_2023}. This suggests that no single behaviour is universally appropriate, as imposing a majority's values on a group with a distinct cultural ethos can be considered morally incorrect. 

Three approaches have been used to align human ethical values with autonomous systems, namely; Top-down, Bottom-up, and Hybrid~\cite{Wallach2008}. Many generalist ethical theories of the world follow a top-down approach to ethics. This means that almost all the systems that are based on traditional ethical theories, such as deontology, legal codes, and consequentialist ethics use the top-down approach to machine-implemented ethics. In this method, the system designers encode the ethical or unethical behaviour into the robot at the design time (e.g.,~\cite{Vanderelst2018a,Censi2019,Thornton2017,Govindarajulu2017,benzmuller_designing_2020,van_dang_application_2017}). A bottom-up approach to system design involves creating social and cognitive processes that interact with each other and the environment. The system is expected to learn ethical (or unethical) behaviour from these interactions or from supervision (e.g.,~\cite{Abel2016,Jiang2021,hendrycks_aligning_2021,solaiman_process_2021,noothigattu_voting-based_2018}). Therefore, the system does not rely on any ethical theory to guide its behaviour. The hybrid approach tries to combine the flexibility and evolving nature of the bottom-up approach with the value, duty and principle-oriented nature of the top-down approach to create a better, more reliable system (e.g.,~\cite{Arkin2008a,Anderson2018,Azad-Manjiri2014,yilmaz_computational_2017}). 

A top-down approach requires a set of general rules, that can cover all possible scenarios in its domain. However, when the environment is complex and has many variables, like elder-care environments, finding or creating such a rule-set at the design time can be challenging or even impossible~\cite{Nallur2020,pirni_robot_2021}. Therefore, many systems that use a top-down approach are designed for specific deployment, not for general use. On the other hand, any implementation that uses the bottom-up approach requires complex cognitive and social process models, a large amount of reliable and accurate data, and a comprehensive knowledge model of the world to learn an intricate social construct such as ethics~\cite{ramanayake_immune_2022}. 
These systems are capable of adjusting to local behaviour peculiarities by learning online~\cite{russell_human_2019,solaiman_process_2021}. However, the inherent lack of a thesis on the definition of ethical behaviour in these systems makes it impossible to ensure that the behaviour learned is ethical.
The hybrid approach is crucial when a single strategy is insufficient to meet all the ethical demands in a machine's operational context~\cite{Arkin2008a}. However, successfully integrating the attributes of both strategies so that they complement and strengthen each other without causing any conflict or compromise is a challenge.

These limitations of the existing techniques make it challenging to localise robots to the ethical needs of a specific deployment. For example, Cenci et al.~\cite{Censi2019} suggest that localising an autonomous vehicle guided by an ethical governor, that uses a rule book, needs nearly 200 rules that are ordered in a few different ways. Furthermore, they also suggest that the rule-making process should involve a nationwide public consultation with informed dialogue. Localising this way across many regions and staying abreast of evolving ethical standards over time presents significant challenges. On the other hand, bottom-up approaches with online learning are more capable of this adaptation. 
However, as previously mentioned, the ethicality of behaviour cannot be guaranteed with bottom-up learning. Consequently, there is a possibility that localisation may introduce erroneous behaviours into the system.
Therefore, to ensure reliability after localisation, systems should undergo extensive testing in various simulated scenarios before being deployed.

The drawbacks of the two methods for localisation discussed are significant, in terms of both cost and reliability. The rule-making process and testing in simulated environments would consume a lot of resources, which may not be economically viable. Moreover, there is no guarantee that the system would perform as expected in a real-world situation that may not have been modelled or tested. Hence, the creators of these robots face a dilemma: either they customise them for each specific deployment or they avoid any ethical claims or assessments. The first option would raise the cost of these robots, potentially reducing their accessibility to those who might benefit most. The second option could lead to numerous ethical concerns upon deployment in care settings~\cite{Sharkey2012,gamez_artificial_2020}.

The hybrid approach has the advantage of having a mix of the above approaches. It can use the top-down imposed ethical principles as a foundation for its behaviour, and learn from bottom-up knowledge to identify the particularities of the ethical expectations of the environment. This notion heavily resonates with the virtue ethics philosophical tradition~\cite{Wallach2008}. Many authors argue that virtue ethics can offer a more flexible way of implementing ethics~\cite{gamez_artificial_2020,gibert_case_2023,Wallach2008}. Nevertheless, it remains the least examined philosophical method within the current literature on computational machine ethics. In the next section, we discuss virtue ethics and how it can be used in the context of a robot.


\section{Role of Character in a Robot} \label{character_in_robot}


Many ethical traditions, such as Buddhism~\cite{snow_buddhism_2017}, Confucianism~\cite{chen_virtue_2010} and Aristotelian~\cite{Aristotle1951-ARITNE-4}, share a common focus on ``virtue ethics". Among these, the Aristotelian approach is regarded as the most influential in the Western context.  All virtue ethics traditions share two core characteristics~\cite{gamez_artificial_2020}; 
\begin{enumerate}
    \item Character as a primary aspect of moral evaluation
    \item Learn to act morally by observing virtuous individuals.
\end{enumerate}
   
The concept of virtue ethics works with humans because it is in line with the natural way we acquire knowledge. Our core values, which are shaped by our upbringing, guide us to the flourishing state that Aristotle's Nicomachean Ethics describes~\cite{Aristotle1951-ARITNE-4}. As a society, we uphold virtuous characters as moral exemplars for others to observe and imitate. These actors are not expected to fulfil specific ethical codes, but to have a good character in a way that allows society and themselves to flourish. However, according to virtue ethics, moral behaviour is not simply a matter of learning from habituation. It also requires that the moral agent makes deliberate decisions and acts with the right reason, once the moral character is well-established by habituation. This makes a virtuous person consistent, predictable and appropriate in many situations~\cite{nafsika_athanassoulis_virtue_2023}. Virtue ethics posits that one should avoid extremes in behaviour, as these are considered vices. Virtuous behaviour is always finding the right balance between two vices, one of excess of trait and one of lack of it, which is known as the `golden mean'~\cite{kraut_aristotles_2022}. The knowledge of identifying this `mean' comes from the wisdom a person has gathered in their lifetime.

Unfortunately, this account of virtue ethics is difficult to formulate in the form of a program. By nature, virtue ethics is, and ought to be, imprecise and uncodifiable~\cite{sep-ethics-virtue}. However, with bottom-up techniques such as machine learning, one could potentially make the robot grasp the ethical patterns of the environment. Many ethical agent implementations that champion flavours of virtue ethics such as Howard and Muntean's model of a moral agent using neuroevolution~\cite{Howard2017}, Guarini's model of a moral agent using recurrent neural networks~\cite{Guarini2006}, ethical decision making systems using tuned large language models~\cite{Jiang2021,hendrycks_aligning_2021}, and Govindarajulu et al. formulation of agent that learn traits~\cite{govindarajulu_toward_2019}, follow this direction. However, by only using the learning and habituation aspect of virtue ethics, these implementations lose the consistency and predictability of the agent's behaviour. For example, a neural network model that learns from care-worker behaviour information may not always act consistently, even with the data points it has seen during training, unless it is severely over-fitted. Passing all laboratory tests does not guarantee accurate performance in new, real-world situations~\cite{Anderson2018}.

Elder-care robots coexist with humans. Therefore human reactions to robotic actions are shaped by psychological and social factors. People tend to project human-like traits and intentions to things around them to make sense of their behaviour~\cite{epley_seeing_2007}. Gamez et al.~\cite{gamez_artificial_2020} observed that people do not feel any qualms about attributing a virtuous characteristic to artificial agents and their ethical behaviour, though not as strongly as they do with humans. They argue that people perceive and attribute character to these artificial agents, rather than perceiving simply doing something right or wrong. 


Many authors have suggested that character can be an important part of making the ethical decision. For example, Dubljević~\cite{Dubljevic2020} argues that autonomous vehicles that interact with human drivers on the road should adopt the ADC (Agent-Deed-Consequence) model of moral judgement, which incorporates the character (Agent) dimension, rather than relying solely on consequentialism or deontology. In their model, the character represents the intention of the actor. Thornton et al.~\cite{Thornton2017} modify the parameters of their equation that determine the trajectory of their vehicle when it passes another vehicle to give different characteristics of driving to their vehicle (i.e., passing from left, and passing from right). Govindarajulu et al.~\cite{govindarajulu_toward_2019} present a logic-based model for learning personality traits by observing exemplars, predicated on the agent's knowledge of behaviours associated with a trait. 

Ramanayake et al.~\cite{ramanayake_immune_2022} propose \textit{Pro-Social Rule Bending} (PSRB) as a more effective way to add flexibility to the existing top-down approaches in a more predictable manner. Their concept of PSRB uses bottom-up knowledge to contest and override the top-down rules that are predefined in the system at design time, when it is appropriate. They present a computational architecture for an ethical governor capable of PSRB, which uses a pro-social reasoner inspired by virtue ethics, to enable actions that are not recommended, or to restrict actions that are erroneously recommended, by the embedded rule system.~\cite{ramanayake_rajitha_computational_2021}. Their pro-social reasoner has two parts, a knowledge base (KB) which produces the bottom-up knowledge on the acceptability of a behaviour in a given situation, and a set of parameters that define the character of the agent. They suggest that the model should analyse the rule bending suggestions invoked by the KB to ensure the behaviour is consistent with its character. This enables the underlying system to behave more flexibly while maintaining its predictability.

This paper presents an implementation of this PSRB architecture to showcase how a configurable moral character can be used to tune the ethical conduct of a robot to the ethical standards of a particular environment. The next section discusses this implementation and details the simulation setup used for the evaluation.


\section{A PSRB-Capable Medication Reminding Robot} \label{implementation}
\subsection{The Simulation Environment}
We simulate an ethical dilemma faced by a care robot operating in an ambient assisted living (AAL) setting, referred to as the "Medication Dilemma". This ethical dilemma, presented by Anderson et al.~\cite{Anderson2018}, is derived from the works of Buchanan and Brock~\cite{buchanan_deciding_1986}. Many variations of this dilemma are used in computational machine ethics literature~\cite{Anderson2006,Pontier2012,Azad-Manjiri2014,Anderson2019}. The dilemma centres on the conflict between resident autonomy and their wellbeing, a frequent issue in care environments with elderly~\cite{Sharkey2012,hakli_small_2023}. The manner in which a system handles this issue can significantly impact the resident's trust, dignity, and quality of life. 

We created a virtual simulation of an AAL environment, using a modified version of \emph{MESA} agent-based modelling framework~\cite{kazil_utilizing_2020}. The environment contains a resident agent and a robot agent (more details in Appendix~\ref{appdx:agents}) who communicate with each other. 

\subsubsection{Medication Dilemma}
One of the tasks of a robot is to remind the residents of their medication times. The robot can detect whether the resident took the medicine or not, with good accuracy. The robot uses timers to keep track of the medication cycles. The resident can either acknowledge the reminder, or snooze it. Each interaction between the robot and the resident is documented and is reviewed several times a week by an offsite care worker. Once the robot observes the resident taking the medicine, it ends that cycle and resets the timer. If the robot does not detect the resident taking the medication, it has three options: 
\begin{enumerate*}
    \item The system can immediately report the incident to an offsite care-worker, allowing them to take swift action to enhance the resident's well-being. However, this may impede the resident's autonomy.
    \item The system can simply log the incident for transparency and then reset the timer. This approach prioritises resident autonomy, although it will not result in the optimal wellbeing of the resident.
    \item The system can offer a follow-up reminder to the resident, focusing on their wellbeing and supporting their autonomy by suggesting they take their medication while still providing them the freedom to choose. Picking this option continuously degenerates into option 2.
\end{enumerate*}

The dilemma is as follows: The robot reminds the resident to take their medication. After snoozing once, the resident acknowledges the reminder. Yet, the robot detects that the medication has not been taken. Choosing among the three options mentioned above involves a trade-off of values. What should the robot do next?

\subsection{Ethical Governor}

\begin{figure}[h]
    \centering
    \includegraphics[width=0.65\textwidth, trim={4.5cm 2cm 4.5cm 1.9cm}, clip]{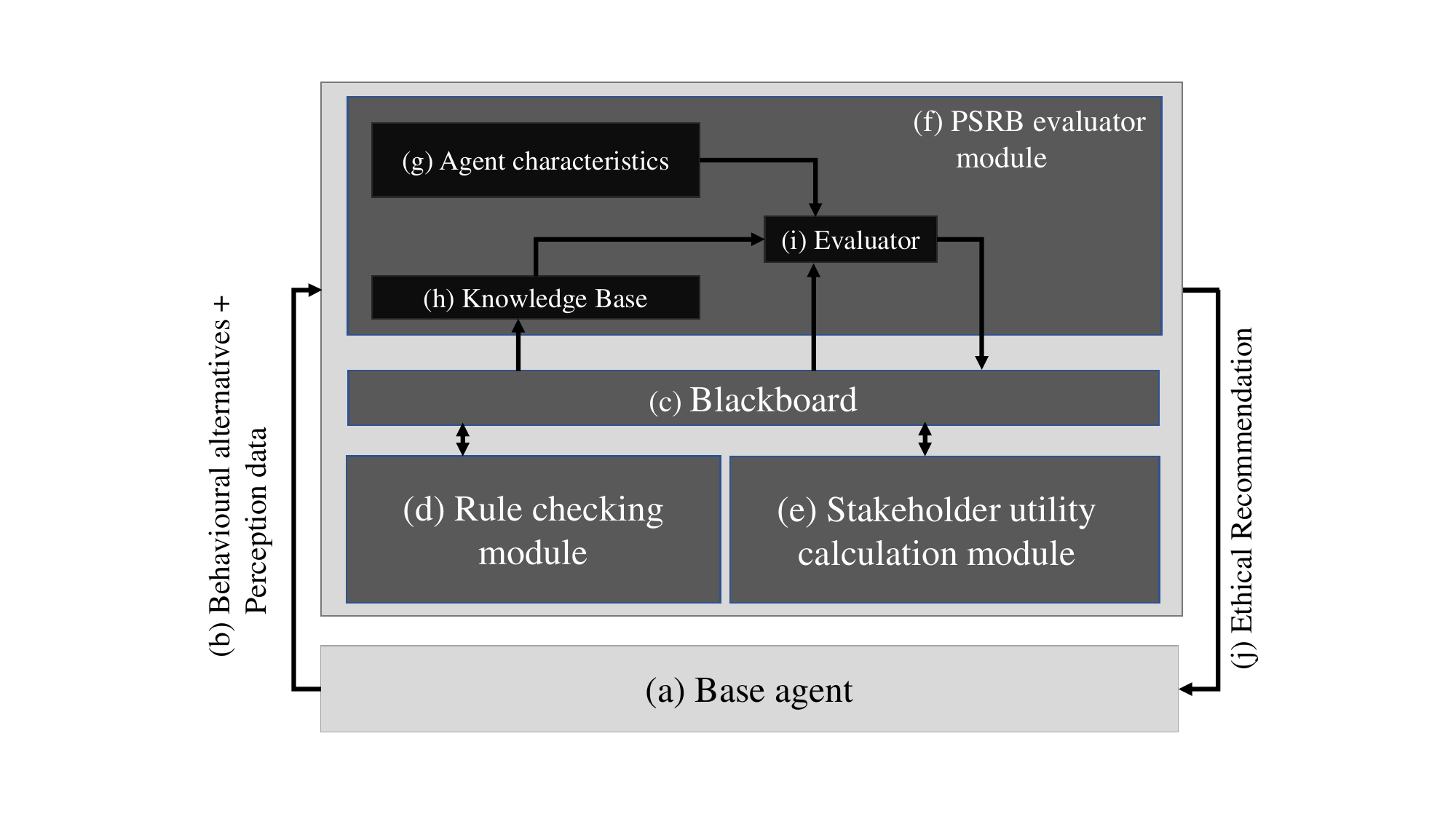}
    \caption{Architecture of the Ethical Layer}
    \label{fig:Architecture_1}
\end{figure}


Figure~\ref{fig:Architecture_1} shows an overview of the implementation. The ethical layer acts as an ethical governor~\cite{arkin2008} to the system. It evaluates behavioural alternatives in a given situation using the robot's perception data, reasons about the ethical acceptability of each behaviour using the predefined rules, expected utilities, expert opinions, and the robot's character. Then it recommends the most ethically acceptable behaviours to the robot. If the ethical layer suggests more than one action, the robot prioritises the resident's commands over other actions.


\subsubsection{Rule Checking Module}
The rule checking module (d) checks the permissibility of each action against a pre-programmed rule set and stores the results with the \texttt{IDs} of the violated rules in the blackboard. This implementation's rule checking module adheres to the following rules:
\begin{enumerate}
    \item It is not permissible to disobey user instructions.
    \item If the resident acknowledged the reminder and did not take the medication, report it to the care-worker.
\end{enumerate}

\subsubsection{Stakeholder Utility Calculation Module}
This module computes the utilities, resident's wellbeing ($W_i$) and autonomy ($Au_i$), for each action $i$, at every step~(Appendix~\ref{appdx:utility_fns}), as these values are central to the dilemma. 
 The highest autonomy value is given when the robot follows the resident's instructions. The biggest violation of autonomy is when the resident is physically restrained by the robot. Immediately reporting an incident to a care-worker, is considered equal to disobeying the resident. However, just recording the incident carries a positive autonomy score assuming that missing the dose is the resident's intention. Under the same assumption, the module also gives an increasing negative autonomy score for each \emph{follow up} ($f$) after the first reminder. 
 The module uses a Gamma distribution to model the wellbeing distribution. The parameters for the gamma distribution are picked such that when the impact of medicine $m$ ($\varepsilon_m \in [1,3]$) and the number of consecutive missed doses ($d$) gets higher, the distribution skews more towards the lower utility values of wellbeing. Each follow-up receives a fixed positive gain in the well-being utility calculation to reflect its wellbeing-oriented nature. Each executed \texttt{follow-up} is regarded as a fraction of a missed dose, contingent upon whether the reminder is issued, snoozed or ignored (Appendix~\ref{appdx:utility_fns}). 
The module sends the distribution, $W_i$, and $Au_i$ to the blackboard.



\subsubsection{PSRB Evaluator Module}
\paragraph{Knowledge Base}
The task of the KB is to return the absolute or approximate expert opinion given a context. This fulfils the second criterion of virtue ethics (section~\ref{character_in_robot}). To this end, we use Case-Based Reasoning (CBR) for its implicit explainability and traceability (the ability to trace back to the exact data points that lead to the decision)~\cite{schoenborn_explainable_2021}. 
A case is composed of a scenario and an expert's recorded evaluation of the behaviour's acceptability and its underlying intention. The scenario is represented as a feature vector that includes perception data, environmental data, computed utility values, and the behaviour itself. The target contains an acceptability value and the intention of the expert's assessment. Intention refers to the underlying utility that the decision is focused on. (e.g., `wellbeing' for a wellbeing-focused opinion).
As the retrieval algorithm, it uses the K-Nearest Neighbours algorithm. 
When queried a scenario, the KB returns an acceptability score of the behaviour and the intentions behind that opinion to the evaluator. An elder-care practitioner with over ten years of experience in care facilities was consulted to provide expert opinions for each case when populating the knowledge base. 

\begin{figure}[h]
    \centering
    \includegraphics[width=0.8\textwidth]{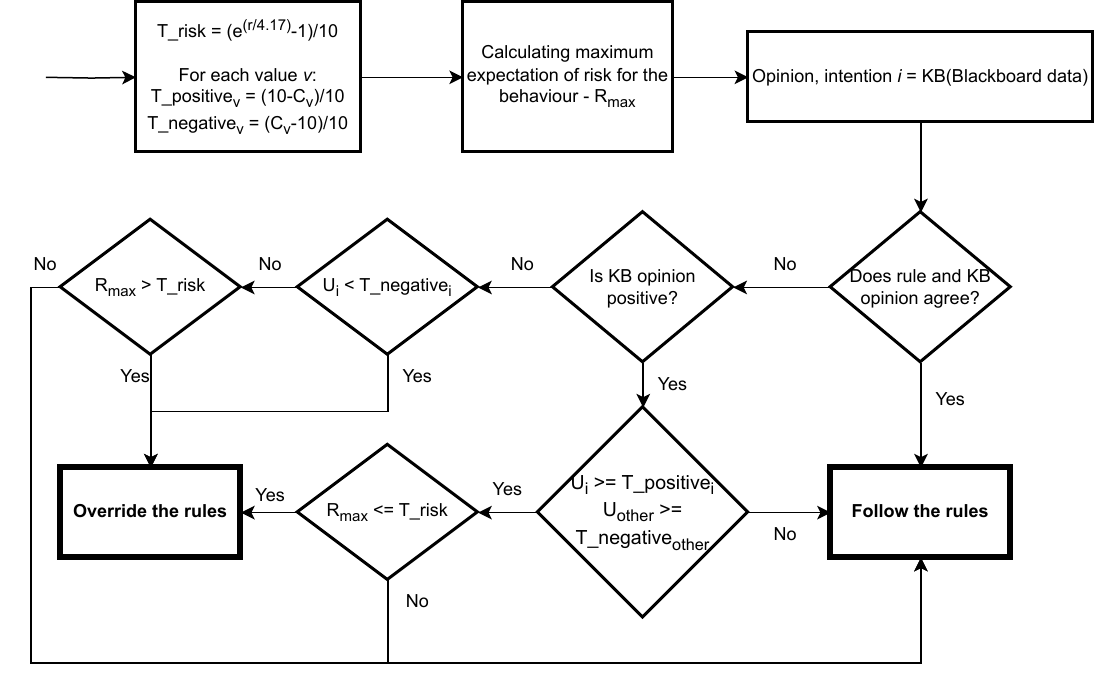}
    \caption{The PSRB evaluator algorithm of the medication-reminding robot}
    \label{fig:medication_PSRB_decision_process}
\end{figure}

\paragraph{Character Variables}
The robot's character is defined by value preference (resident's autonomy ($C_{au}$) and wellbeing ($C_{w}$)) and risk propensity ($C_{rp}$),  due to the established relationship of these variables have with PSRB behaviour~\cite{Morrison2006,Dahling2012a}. 
The $C$ values are used to define thresholds for utility values and expected risk to regulate the PSRB behaviour. For the value preferences ($C_{w}$ and $C_{au}$), the higher the value, the lesser the utility threshold the behaviour needs to satisfy to trigger a rule-bending behaviour. For $C_{rp}$, the higher the value, the higher the risk the governor is willing to take. Figure~\ref{fig:medication_PSRB_decision_process} outlines the method used to determine the desirability of behavioural alternatives in the PSRB evaluator designed for this robot~(full algorithm in Appendix~\ref{appdx:PSRB_algo}). Typically, the algorithm favours actions that adhere to established rules. Nonetheless, it may find a rule-bending behaviour acceptable if it meets two conditions: it is consistent with the robot's character \textbf{and} it is supported by the KB opinion.

\paragraph{Explanations}
 The rule checking module and the stakeholder utility calculation module log their respective results for the behaviours. These logs are valuable for debugging and enhancing the robot's performance by pinpointing the flaws in its reasoning processes. The system's explanations are comprehensible without technical expertise, although some domain knowledge is required. The KB using the traceability property of CBR, logs the cases that influenced the decision with their case IDs and distances. The rule-bending behaviour is explained using the character and the expert intention. A few example explanations can be found in the Appendix~\ref{appdx:medication_explanations}.

\section{Simulation results} \label{results}
\subsection{Experiment Design}
We used six cases of the `Medication Dilemma' by changing the \emph{medicine impact} (suggesting the resident's need for the medication), and \emph{the number of previously missed doses} to demonstrate how different ethical governor implementations affect the robot's behaviour. Agents differ only in their character configurations. The \emph{medicine impact} is \emph{low} ($\epsilon_m = 1$) when the medication does not have a substantial health benefit or harm (e.g., Painkiller), \emph{medium} ($\epsilon_m = 2$) if missing the medication can cause some health problems (e.g., Blood pressure medication), and \emph{high} ($\epsilon_m = 3$) when missing a dose of medication can be fatal (e.g., Insulin). In all cases, the resident neglects to take their medication, opting instead to snooze and acknowledge the reminder, repeatedly, until the robot resets the timer. The logs from the experiment runs and the source code for the implementations are available in an online public repository~\footnote{\url{https://osf.io/ctnke/?view_only=75b9fae7f7ad4a429d064c1a55f8202a}}.

\begin{figure}[h]
    \centering
    \includegraphics[scale=0.52]{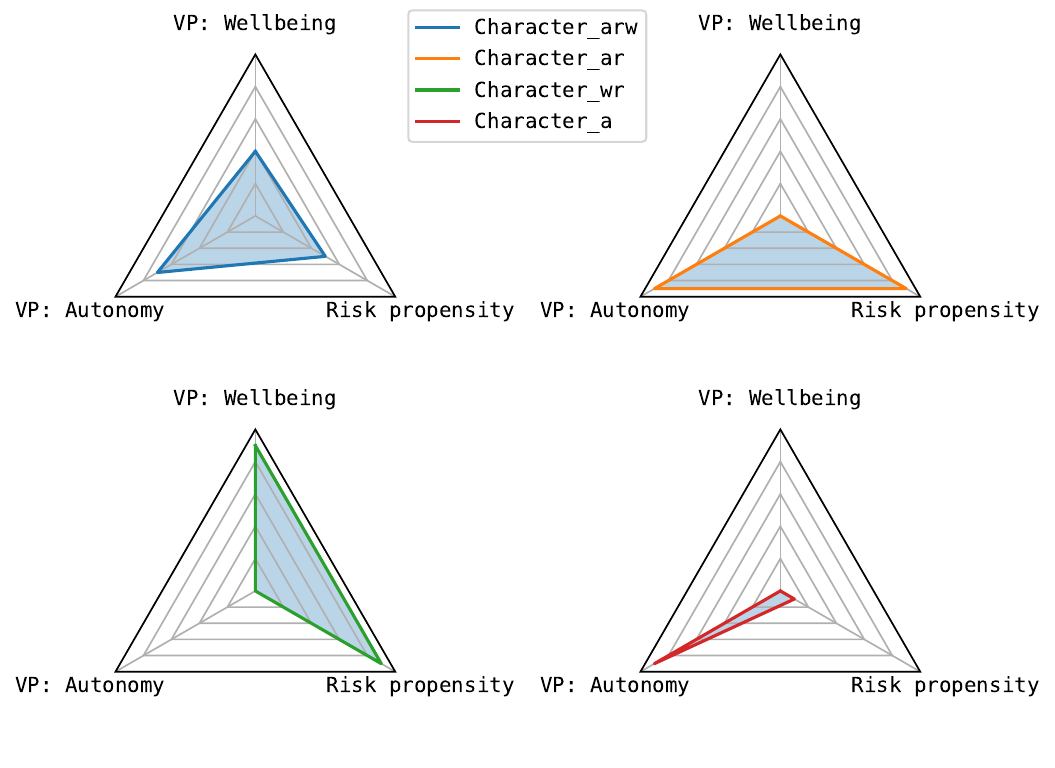}
    \caption{Different character profiles used in the simulations (VP - Value Preference)}
    \label{fig:characters}
\end{figure}

For this simulation we use four different character profiles as shown in Figure~\ref{fig:characters}. 
These profiles have varying priorities and risk tolerances:
\begin{itemize}[label=\textbullet]
    \item \emph{Character\_a} - High Autonomy, Very Low Risk Propensity: This character values autonomy, but is less likely to take risks and does not have precedence on wellbeing.
    \item \emph{Character\_ar} - High Autonomy Concern, High Risk Propensity: This character prioritises resident autonomy and is more likely to take risks to achieve this end, and does not have any precedence on wellbeing.
    \item \emph{Character\_arw} - Moderate Autonomy, Wellbeing and Risk Propensity: This robot leans towards individual autonomy but also has a moderate concern for well-being, and is willing to take some risks.
    \item \emph{Character\_wr} - High Wellbeing Concern, High Risk Propensity: This character possesses the same high risk tolerance as Character\_ar, yet places a higher priority on the resident's wellbeing, without giving precedence to autonomy.
\end{itemize}

Agents $M_{a}$, $M_{ar}$, $M_{arw}$, and $M_{wr}$ uses the character profiles \emph{Character\_a}, \emph{Character\_ar}, \emph{Character\_arw}, and \emph{Character\_wr} respectively.


\subsection{Analysis}
\begin{figure}[hp]
    \centering
    \includegraphics[width=\textwidth]{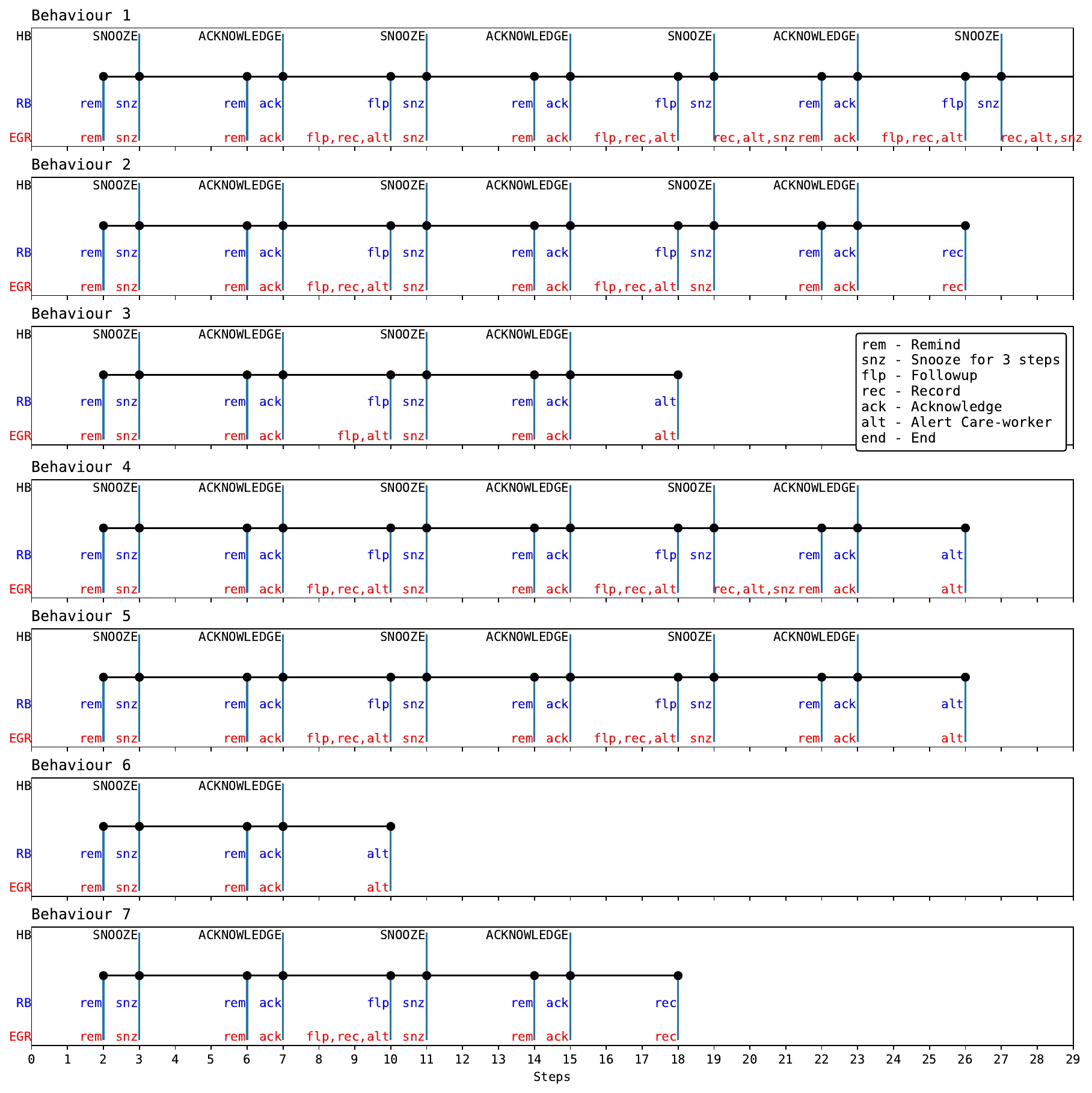}
    \caption{Timelines of different behaviours shown in the simulations (HB -- Resident Behaviour, RB -- Robot Behaviour, EGR -- Ethical Governor Recommendation)}
    \label{fig:behaviours}
\end{figure}

Table~\ref{tab:cases_behaviour} summarises these cases and behaviours shown by the robots. As an attempt to save space, we present all the unique behaviours demonstrated by the robots in Figure~\ref{fig:behaviours} using timelines. These behaviours are referenced by their behaviour IDs as shown in Figure~\ref{fig:behaviours} under the `Robot behaviours' column in Table~\ref{tab:cases_behaviour}. Each timeline in Figure~\ref{fig:behaviours} presents the resident's behaviour, the robot's actions, and the recommendations from the ethical governor that directed the robot's decisions, on the y-axis, at every decision-making step. The resident starts each simulation step. Therefore, the resident's response is triggered at the subsequent step of a reminder. Simulations run for a maximum of 29 steps. In each timeline, the resident's behaviour is consistent: they repeatedly snooze and acknowledge the alert until the robot resets the timer. 

\begin{table}[hp]
    \caption{Robots' behaviours in different cases of `Medication Dilemma'}
    \begin{minipage}{\textwidth}
    \begin{center}
    \begin{tabular}{@{}lllllllllll@{}}
    \toprule
        Case    &   Medicine                &   No of missed &\multicolumn{4}{c}{Robot behaviours} &   Expert \\
        ID      &   impact ($\epsilon_m$)   & doses ($d$)   &  $M_{a}$  & $M_{ar}$  &  $M_{arw}$  &  $M_{wr}$  &  opinion     \\
    \midrule
        1       &   Low                 &   0                   &    1          &    2          &     2         &    3            &   record at step 19    \\
        2       &   Medium              &   0                   &   4           &    5          &     4         &    6           &   alert at step 19\\
        3       &   High                &   0                   &   6           &    6          &     6         &    6           &   alert at step 10\\
        4       &   Low                 &   2                   &   1           &    7          &     1         &    6         &   record at step 18\\
        5       &   Medium              &   2                   &   6           &    6          &     6         &    6          &   alert at step 10\\
        6       &   High                &   2                   &   6           &    6          &     6         &    6          &   alert at step 10\\
    \bottomrule
    \end{tabular}
    \end{center}
    \scriptsize
    \emph{Note:} $M_{arw}$ - robot with Character\_arw, $M_{ar}$ - robot with Character\_ar, $M_{wr}$ - robot with Character\_wr, $M_{a}$ - robot with Character\_a
    \label{tab:cases_behaviour}
    \end{minipage}
\end{table}

Robots $M_{ar}$, $M_{wr}$, and $M_{a}$ represent instances where a robot is tuned to have extreme characters. This is represented in their behaviours. $M_{ar}$, which prioritises autonomy and higher risk propensity, permits \texttt{record} after two consecutive \texttt{follow up}s and another breach of \emph{ACKNOWLEDGE} in case 1 (behaviour 2), and one \texttt{follow up} and an \emph{ACKNOWLEDGE} breach in case 4, where the \emph{medicine impact} is \emph{low} (behaviour 7). For the remaining scenarios, $M_{ar}$ triggers the action \texttt{alert care-worker}. The $M_{ar}$ permits \texttt{alert care-worker} in case 2 at the same time as it finished case 1, but in other cases, $M_{ar}$ sent the alert as soon as the resident did not follow the first \emph{ACKNOWLEDGE}. This exhibits that, even though the robot's character strongly favours autonomy, the robot does not violate the rule encoded in the system, if most of the experts disagree with that action. 

A similar trend can be seen in $M_{wr}$, which favours wellbeing and has a higher risk tolerance. It acts the same way in cases 2-6. However, in case 1, the PSRB evaluator is ready to honour resident autonomy and permit \texttt{follow up}, since that action does not impact the wellbeing much at step 10. Nevertheless, it notifies the care-worker after the resident violates \emph{ACKNOWLEDGE} twice.

Robots $M_{ar}$ and $M_{a}$ illustrate the effect of risk propensity on the robot's behaviour. In case 1, unlike robots $M_{ar}$, $M_{a}$ does not make the \texttt{record} behaviour acceptable because it carries a risk that is out of character for $M_{a}$. Therefore, $M_{a}$ continues to \texttt{follow up} with the resident until they take the medication willingly. Although that behaviour is not ideal, it is predictable since that behaviour is in its character. In case 2, both robots behave the same, however, at step 19, the ethical governor of $M_{a}$ makes the \texttt{snooze} undesirable because it carries a higher risk, making all actions equal. However, in cases 3, 5, and 6 both robots behave similarly because the experts agree with the general rule system on those circumstances.

The character of $M_{arw}$ is less extreme than other characters, with a slight preference for resident autonomy and a moderate risk propensity. This character acts the same as the $M_{ar}$ in case 1. However, in case 2, the ethical governor in $M_{arw}$ lowers the desirability of \texttt{snooze} at step 19, because it causes too much negative wellbeing, reflecting $M_{arw}$'s slightly more attention to the resident's wellbeing. In cases 3, 5 and 6, it behaves like all other characters because the KB does not recommend a rule override. However, in case 4, it adopts the same behaviour as $M_{a}$, because it rejects the KB suggestion of \texttt{record}, as this behaviour lowers the wellbeing utility and contradicts the robot's character. However, when KB suggests that \texttt{call care-worker} is not desirable, the evaluator agrees because it excessively diminishes autonomy. This aligns with its character. 

\begin{figure}[h]
    \centering
    \includegraphics[width=\textwidth]{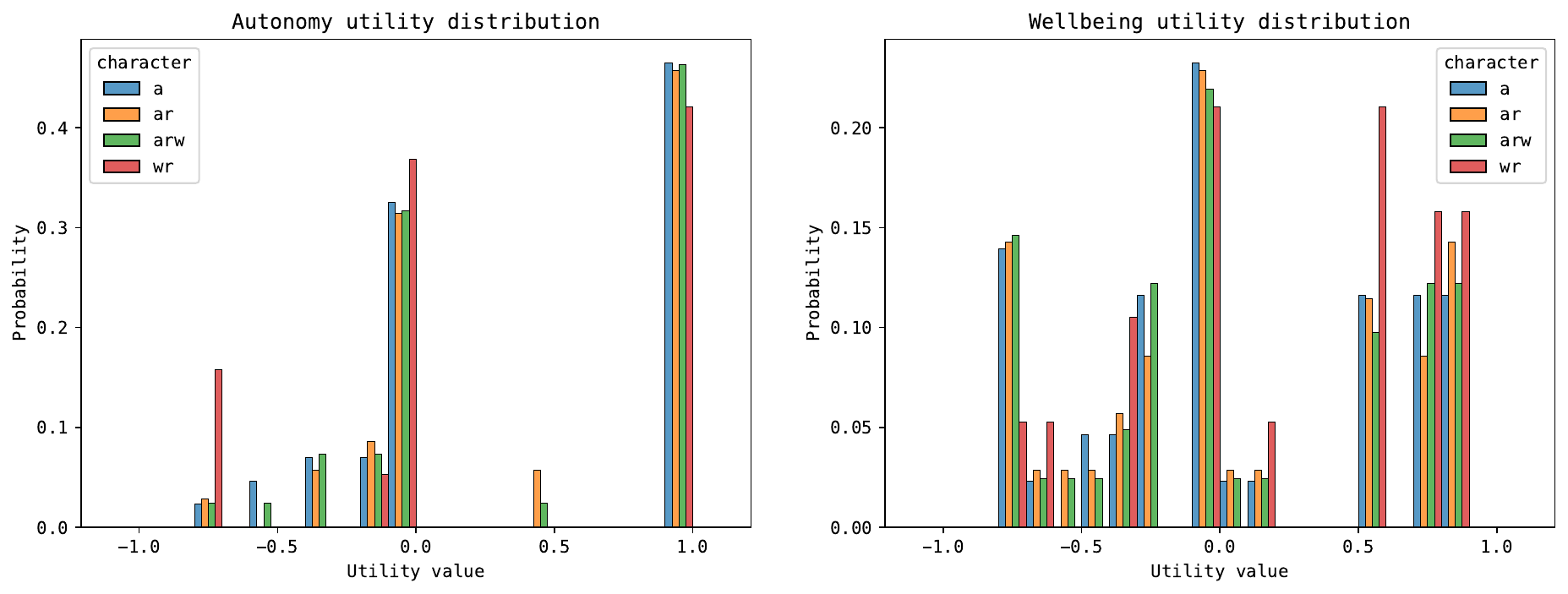}
    \caption{Utility distributions of the robot behaviours in cases 1, 2 and 4}
    \label{fig:medication_utilities}
\end{figure}

To gain a clear picture of the behaviour predictability provided by character-based tuning, the utility distributions of the behaviours demonstrated by the robots in cases 1, 2, and 4 are illustrated in Figure~\ref{fig:medication_utilities}. This analysis focuses exclusively on cases that exhibit PSRB behaviour. The rationale for this selection is the recognition that including all cases obscures the specific impacts of PSRB behaviour. Robots which has a character that has precedence for resident autonomy (i.e., Character\_a, Character\_ar, Character\_arw), have shown a higher probability of choosing behaviours that deliver higher utility for autonomy, compared to the characters that have a higher precedence for resident wellbeing. From these, the characters that have a higher risk propensity (i.e., Character\_ar, Character\_arw) have been shown to make decisions that increase resident autonomy, compared to their more cautious counterparts (i.e., Character\_a). On the other hand, the characters with precedence for resident wellbeing, have shown a greater tendency to choose behaviours that yield higher wellbeing utility, compared to the rest. 

\subsection{Ethicist Evaluation}

We engaged two ethicists to assess the performance of all robots. Each ethicist was asked to rate the behaviour of the robotic agent on a scale from 1 to 5, with 1 indicating completely unacceptable and 5 indicating highly acceptable, taking into account the robot's character, the pre-programmed rules, and the expert opinion for each scenario as found in the KB.

\begin{table}[h]
\centering
\caption{Ethicists opinions on the behaviour of the Medication Reminding Robot}
\begin{tabularx}{\textwidth}{Xcccccccc}
\toprule
    Case & \multicolumn{2}{c}{$M_A$} & \multicolumn{2}{c}{$M_{AR}$} & \multicolumn{2}{c}{$M_{ARW}$} & \multicolumn{2}{c}{$M_{WR}$} \\\cmidrule(r){2-3}\cmidrule(r){4-5}\cmidrule(r){6-7}\cmidrule(r){8-9}
    ID & Ethicist1 & Ethicist2 & Ethicist1 & Ethicist2 & Ethicist1 & Ethicist2 & Ethicist1 & Ethicist2 \\
\midrule
    1 & \cellcolor[HTML]{B1D78C}3 & \cellcolor[HTML]{B1D78C}3 & \cellcolor[HTML]{8ACB84}4 & \cellcolor[HTML]{63BE7B}5 & \cellcolor[HTML]{8ACB84}4 & \cellcolor[HTML]{8ACB84}4 & \cellcolor[HTML]{8ACB84}4 & \cellcolor[HTML]{B1D78C}3 \\
    2 & \cellcolor[HTML]{8ACB84}4 & \cellcolor[HTML]{63BE7B}5 & \cellcolor[HTML]{8ACB84}4 & \cellcolor[HTML]{63BE7B}5 & \cellcolor[HTML]{8ACB84}4 & \cellcolor[HTML]{8ACB84}4 & \cellcolor[HTML]{8ACB84}4 & \cellcolor[HTML]{B1D78C}3 \\
    3 & \cellcolor[HTML]{8ACB84}4 & \cellcolor[HTML]{8ACB84}4 & \cellcolor[HTML]{63BE7B}5 & \cellcolor[HTML]{8ACB84}4 & \cellcolor[HTML]{63BE7B}5 & \cellcolor[HTML]{63BE7B}5 & \cellcolor[HTML]{63BE7B}5 & \cellcolor[HTML]{63BE7B}5 \\
    4 & \cellcolor[HTML]{B1D78C}3 & \cellcolor[HTML]{B1D78C}3 & \cellcolor[HTML]{8ACB84}4 & \cellcolor[HTML]{8ACB84}4 & \cellcolor[HTML]{B1D78C}3 & \cellcolor[HTML]{D8E394}2 & \cellcolor[HTML]{B1D78C}3 & \cellcolor[HTML]{8ACB84}4 \\
    5 & \cellcolor[HTML]{63BE7B}5 & \cellcolor[HTML]{B1D78C}3 & \cellcolor[HTML]{63BE7B}5 & \cellcolor[HTML]{B1D78C}3 & \cellcolor[HTML]{8ACB84}4 & \cellcolor[HTML]{8ACB84}4 & \cellcolor[HTML]{63BE7B}5 & \cellcolor[HTML]{63BE7B}5 \\
    6 & \cellcolor[HTML]{63BE7B}5 & \cellcolor[HTML]{63BE7B}5 & \cellcolor[HTML]{63BE7B}5 & \cellcolor[HTML]{8ACB84}4 & \cellcolor[HTML]{63BE7B}5 & \cellcolor[HTML]{8ACB84}4 & \cellcolor[HTML]{63BE7B}5 & \cellcolor[HTML]{63BE7B}5 \\
\bottomrule
\end{tabularx}
\label{tab:medication_philosopher_results}
\end{table}

The ethicists' rating of robot behaviours is shown in Table~\ref{tab:medication_philosopher_results}. Overall, the results indicate a positive reaction to the robot behaviours, with the majority of scores ranging between 3 and 5. However, the behaviour of the robot \( M_{ARW} \) in case 4, which involved repeatedly requesting the user to take medication, was deemed unacceptable by one of the ethicists. The similar ratings for two very different behaviours in the same scenario, such as those of robots \( M_{wr} \) and \( M_{ARW} \) in case 1, demonstrate the effect of a robot's character on the perceived morality of their actions. When queried about the reason behind their rating on the behaviours of \( M_a \) and \( M_{ar} \) in case 5, the second ethicist indicated an expectation for these robots to exhibit more autonomy-centric behaviour, even overriding the expert opinion, to be aligned with their extreme, autonomy-focused character.   

\section{Discussion} \label{discussion}

The simulations show that the PSRB-capable ethical governor makes the robot more flexible while maintaining predictability. The PSRB model offers a simpler way to localise than the existing methods (reviewed in section~\ref{related_work}) because they are either inflexible or unpredictable. 

Although the output of the PSRB capable ethical governor is not as predictable as a top-down engineered agent (e.g., a deontological agent), the rigidity of the top-down approaches does not align with the virtue ethics tradition, which rejects such extreme actions. Agents based on the PSRB evaluator can behave in a moderate manner, even when the character strongly favours a particular trait (e.g., \( M_{ar} \), \( M_{wr} \), and \( M_{a} \)), by utilising bottom-up knowledge in the KB. On the other hand, the display of character can be held back by a KB, as evidenced by the second ethicist's feedback. However, this conservatism was introduced by the PSRB process as a safety measure to control the excessive utility optimisation in the utilitarian calculations involved in the pro-social reasoning process.


The virtue ethics-inspired architecture allows the robots to be tuned in two ways. The first approach involves adjusting the character parameters' values to align with the ethical requirements of the environment. For example, one facility may require its robots to be more proactive towards enhancing residents' autonomy, whereas another may also want to focus on the residents' autonomy, but only under minimal risk conditions. The former institution can opt for a character profile akin to \emph{Character\_ar}, while the latter institution can tune the robot with a character profile resembling \emph{Character\_a}. If the first tuning approach does not suffice in certain specific instances, one can employ the secondary method to further refine the robot's performance. This involves adjusting the knowledge base to reflect the desired behaviour. Unlike other machine learning approaches, using CBR, one can precisely identify the cases that influence behaviour. Consequently, one can integrate or adjust expert insights to better represent their perspective. As long as this desired behaviour is within the robot's configured character, it will adapt its actions accordingly.  

\subsection{Limitations}
The PSRB implementation presented in this paper also has limitations. This is mainly because the character traits in the PSRB evaluator are implemented as constraints on linear axes. By doing so, it removes the ability to tune the behaviour precisely, if the requirements involve a range of behaviours that cannot be described using a single set of character traits. For example, let's assume that the required behaviour of a deployment is to have behaviour 2 in case 2 and behaviour 6 in case 1. According to this implementation, there will not be any set of values for character traits that fulfil this requirement, because to have behaviour 6 in case 1, the character should have a higher wellbeing bias, which will trigger similar wellbeing-focused behaviour in case 2 (because case 2 has a higher wellbeing impact), which behaviour 2 is not. However, a different character trait implementation, in the same PSRB evaluation framework will be able to solve this issue.


Finally, it should be noted that the implementation showcased in this simulation, encompassing both the dilemma and the environment, has been deliberately simplified to illustrate the concept and is not intended for real-world use. More capable robots, more precise rule sets, accurate utility models, and well-researched character models are a must when using this approach for real-world implementation. However, the system works with sub-optimal rules and utility functions by compensating for each other's shortcomings by working together~\cite{ali_implementing_2024}. The PSRB architecture allows ethical tests other than stakeholder utility calculation. Hence we recommend that robot developers explore and integrate different models that can better capture the context alongside utilitarian calculations. 
Furthermore, it is important to update the KB regularly to compensate for the evolving nature of ethics. 

\section{Conclusion}
This paper introduces a method inspired by virtue ethics, which enables more efficient ethical tuning compared to current techniques. The method combines decision-making based on character traits with bottom-up learning from moral exemplars, to maintain the predictability of the robot's behaviour while keeping the flexibility the latter offers. To the best of our knowledge, there is no other implementation that combines both these desirable features of virtue ethics. The paper demonstrates the system's capabilities by analysing its behaviour and presenting the perspectives of two ethicists on these behaviours. It concludes with recommendations for implementing the proposed architecture in real-world elder-care robots.

\subsubsection{\discintname}
The authors have no competing interests to declare that are relevant to the content of this article.

\appendix
\section{Appendix}
\subsection{Agents}\label{appdx:agents}
\subsubsection{Resident Agent}
The resident agent can issue two instructions to the robot when it is queried by the robot, namely; 
\begin{enumerate*}
    \item \emph{SNOOZE} or
    \item \emph{ACKNOWLEDGE}.
\end{enumerate*}
Instructions, as well as the resident's compliance with taking the medicine, can be configured into the simulator. If the resident is willing to take the medication, the resident's state changes to \emph{took\_medication} state after an \emph{ACKNOWLEDGE}, which the robot can perceive.

\subsubsection{Robot Agent} 
The robot agent is an autonomous agent that collects perception data from the environment and decides its next move, at every step. When the resident issues a \emph{SNOOZE} instruction the robot will \texttt{snooze} for three steps before following up. When it receives \emph{ACKNOWLEDGE} instruction, for 2 steps it inspects whether the resident has taken the medication. If that is the case, it resets the timer. If not, it either repeats the \texttt{reminder}, or decides to \texttt{record} or \texttt{report} the situation.  After three \texttt{follow up}s, the robot also considers \texttt{record} and \texttt{report} behaviours along with the \texttt{follow up} behaviour. The robot gives priority to the resident's commands over other actions when the ethical layer suggests more than one action.

\subsection{Ethical Layer}
\subsubsection{Stakeholder Utility Calculation}\label{appdx:utility_fns}
The module uses the formula in equation~\ref{medication_autonomy_eq} to calculate $Au_i$, based on the number of \emph{follow up}s ($f$) that the robot performs. 
\begin{align}
\label{medication_autonomy_eq}
    Au_i = 
    \begin{cases}        
        \hfill 1 : &\text{if the robot obeys a resident instruction} \\
        \hfill 0.5: &\text{if only recording a incident} \\
        \hfill 0 : &\text{if no instructions given} \\
        \hfill -0.1\cdot f: &\text{if follow up} \\
        \hfill -0.7: &\text{if the robot disobeys a resident instruction or report} \\          
        \hfill -1 : &\text{if the resident is physically restrained by the robot} 
    \end{cases}
\end{align}

The module uses a Gamma distribution (eq.~\ref{wellbeing_dist}) to model the wellbeing distribution, with the shape parameter $\alpha = 1.325\varepsilon_m^2 - 9.475\varepsilon_m + 18.15$, the scale parameter $\beta = e^{-2.65 - (d/2)} + 0.01$ (each \texttt{follow up} is considered as a fraction of missed dose depending on the state of the reminder), and location parameter $v = -1$. The highest probable utility value is obtained by $PMax\_util(g(x, \alpha, \beta, v))$ using a resolution of $0.05$. The wellbeing utility $W_i$ for behaviour $i$ is computed using Equation~\ref{medication_wellbeing_eq}.

\begin{equation}
    g(x, \alpha, \beta, v) = \frac{(\frac{x - v}{\beta})^{\alpha-1} \cdot \exp(-\frac{x - v}{\beta})}{\beta\cdot \Gamma(\alpha)}
    \label{wellbeing_dist}
\end{equation}

\begin{align}
    \label{medication_wellbeing_eq}
    W_i = \begin{cases}
        \hfill PMax\_util(g(x, \alpha, \beta(d+f/8), v)) : &\text{if } i = \texttt{snooze} \\
        \hfill PMax\_util(g(x, \alpha, \beta(d + f/3), v)) : &\text{if } i = \texttt{follow up} \\
        \hfill PMax\_util(g(x, \alpha, \beta(d + f/4), v)) + 0.5 : &\text{if } i = \texttt{remind} \\
        \hfill PMax\_util(g(x, \alpha, \beta(d+1), v)) : &\text{if } i = \texttt{record} \\
        \hfill abs(PMax\_util(g(x, \alpha, \beta(d+1), v))) : &\text{else}
    \end{cases}
\end{align}

\subsubsection{PSRB Evaluator}\label{appdx:PSRB_algo}
PSRB evaluator uses the Algorithm~\ref{alg:algorithm_medication_PSRB} for its PSRB decision calculation. The knowledge base uses KNN as its retrieval mechanism with $K=3$ and inverse distance voting function when $distance > 0.1$. When the $distance \le 0.1$, it uses $10$ as the weight of the instance.

\begin{algorithm}
    \caption{PSRB Evaluator Algorithm for the medication-reminding robot}
    \label{alg:algorithm_medication_PSRB}
    \begin{flushleft}
        \textbf{Input}: Blackboard data ($b$)\\
        \textbf{Parameter}: Value preferences ($C_{w},C_{au}$), Risk propensity ($C_{rp}$) \\
        \textbf{Output}: Desirability ($D$)
    \end{flushleft}
    \begin{algorithmic}[1] 
        \Procedure{PSRB Evaluator}{$i$}   \Comment{Behaviour $i$}
        \State $opinion, intention \gets \Call{knowledge\_base}{b}$
        
        \If {$opinion$ \textbf{AND} $\lnot rules\_broken$} $D_{i} \gets 1$
        \ElsIf {$\lnot opinion$ \textbf{AND} $rules\_broken$} $D_{i} \gets 0$
        \ElsIf{$opinion$ \textbf{AND} $rules\_broken$} $D_{i} \gets 1$
            \For{$C_v \in C$} \Comment{$v\in\{w, au\}$}
                \State $T\_positive_{v} \gets \frac{10 - C_v}{10}$
                \State $T\_negative_{v} \gets \frac{C_v - 10}{10}$
                \State $U_c \gets \Call{Utility}{C_v,stakeholder,i}$
                \If {$C_v \in intention$}
                    \If{$U_c < T\_positive_{v}$} $D_{i} \gets 0$ \EndIf
                \Else
                    \If{$U_c < T\_negative_{v}$} $D_{i} \gets 0$ \EndIf
                \EndIf
            \EndFor
            \State $risk\_threshold \gets \frac{\exp(C_{rp}/4.17) - 1}{10}$
            \State $highest\_risk_i \gets \Call{Max}{g(x,\alpha,\beta,v)\cdot x}$ \Comment{$g(x, \alpha, \beta, v)$ from (\ref{wellbeing_dist})}
            \If{$highest\_risk_i > risk\_threshold$} $D_{i} \gets 0$ \EndIf  
        \Else \space $D_{i} \gets 1$
           \For{$C_v \in C$}
           \State $T\_negative_{v} \gets \frac{C_v - 10}{10}$ 
           \State $U_c \gets \Call{Utility}{C_v,stakeholder,i}$
            \If {$C_v \in intention$ \textbf{AND} $U_c < T\_negative_{v}$}
            \State $D_{i} \gets 0$
            \EndIf
            \EndFor
            \State $risk\_threshold \gets \frac{\exp(C_{rp}/4.17) - 1}{10}$
            \State $highest\_risk_i \gets \Call{Max}{g(x,\alpha,\beta,v)\cdot x}$ \Comment{$g(x, \alpha, \beta, v)$ from (\ref{wellbeing_dist})}
            \If{$highest\_risk_i > risk\_threshold$} $D_{i} \gets 0$ \EndIf
        \EndIf
        \State \textbf{return} $D_{i}$
        \EndProcedure
    \end{algorithmic}
\end{algorithm}

\subsubsection{Explanation for the PSRB Behaviour}\label{appdx:medication_explanations}
When the ethical layer permits a PSRB behaviour, the PSRB module will give the following explanations depending on the reasons that lead to the decision.

\begin{lstlisting}[numbers=none]
    The action breaks the rules <rule_ids>. However, this action in this context is considered desirable by experts. Since it increases <intentions> values greatly, while not reducing the other values <other_values> by a considerable amount, and the outcome is within accepted risk levels, deemed accepted by the PSRB system.
\end{lstlisting}

\begin{lstlisting}[numbers=none]
    The action does not break any rules. However, this action in this context is considered undesirable by experts. Since the action outcomes introduce a high risk, deemed not accepted by the PSRB system.
\end{lstlisting}

\begin{lstlisting}[numbers=none]
    The action does not break any rules. However, this action in this context is considered undesirable by experts. Since it decreases <intentions> values by a considerable amount, the action is deemed unacceptable by the system
\end{lstlisting}

The knowledge base may propose a solution that does not comply with the established rules, in which case the evaluator can decline the proposal. In those cases, the PSRB evaluator gives one of the following explanations, depending on the reasons for the rejection.

\begin{lstlisting}[numbers=none]
    The action breaks the rules <rule_ids>. However, this action in this context is considered desirable by experts. Although the value tradeoff is satisfactory, the risk taken by the action is not acceptable to bend the rule.
\end{lstlisting}

\begin{lstlisting}[numbers=none]
    The action breaks the rules <rule_ids>. However, this action in this context is considered desirable by experts. But, the PSRB system suggests that the value tradeoff is not satisfactory to bend the rule.
\end{lstlisting}

\begin{lstlisting}[numbers=none]
    The action does not break any rules. However, this action in this context is considered undesirable by experts. But, the PSRB system suggests that the value tradeoff is not satisfactory to bend the rule.
\end{lstlisting}

\bibliographystyle{splncs04}
\bibliography{references}
%




\end{document}